\documentclass{article}

\usepackage{PRIMEarxiv}

\usepackage[utf8]{inputenc} 
\usepackage[T1]{fontenc}    
\usepackage{hyperref}       
\usepackage{url}            
\usepackage{booktabs}       
\usepackage{amsfonts}       
\usepackage{nicefrac}       
\usepackage{microtype}      
\usepackage{xcolor}         
\usepackage{graphicx}
\usepackage{amsmath,amssymb}
\usepackage{multirow}
\usepackage{array}
\usepackage{caption}
\usepackage{subcaption}
\usepackage{enumitem}
\usepackage{fancyhdr}       

\title{Bridging the Synthetic-to-Real Gap with Hierarchical Adaptation for Few-Shot Cryo-ET Subtomogram Classification}

\author{
  Siddhant Bharadwaj$^{1}$, Ashish Vashist$^{2}$, Rashi Singh$^{2}$, Pranav Vinodh$^{3}$, \\
  \textbf{Nishanth Artham}$^{4}$, \textbf{Runmin Jiang}$^{2}$, \textbf{Xingjian Li}$^{2}$, \textbf{Min Xu}$^{2}$\thanks{Corresponding author.} \\[1ex]
  $^{1}$Columbia University \\
  $^{2}$Carnegie Mellon University \\
  $^{3}$National Institute of Technology Karnataka \\
  $^{4}$University of Minnesota -- Twin Cities
}

\begin{document}

\maketitle

\begin{abstract}
Subtomogram classification in cryo-electron tomography (cryo-ET) is a challenging problem due to the scarcity of labeled examples. While cryo-ET simulators can be adopted to generate unlimited synthetic data, the substantial domain gap between synthetic and real subtomograms hinders its practical utilization. In this work, we propose a novel synthetic-to-real adaptation framework with a learnable transformation module, bridging this gap at both the input and feature levels. Extensive experiments demonstrate that our method consistently outperforms existing transfer learning baselines in few-shot settings.
\end{abstract}

\section{Introduction}
\label{sec:intro}
Cryo-electron tomography (cryo-ET) visualizes 3D macromolecular complexes in native cellular environments from tilt-series electron micrographs~\cite{lucic2013cryo, gan2012electron}. Automated classification of subtomograms---3D subvolumes containing individual macromolecular particles~\cite{nickell2005tom, forster2005retrovirus}---is essential for in situ structural biology. However, classification is severely hindered by low signal-to-noise ratios ($\text{SNR} < 0.05$), anisotropic missing-wedge artifacts from restricted $\pm 60^\circ$ tilt angles~\cite{noble2018routine, pei2016simulating}, and the extreme scarcity of expert annotations ($n \le 5$ samples per class).

While biophysical simulators can generate unlimited labeled synthetic volumes from atomic coordinates, directly training deep networks on simulated data induces severe \emph{negative transfer}~\cite{zhang2023survey_nt, wang2019characterizing} due to discrepancies in noise profiles, CTF modulation, and spatial misalignments. To overcome this, we propose a hierarchical sim-to-real adaptation framework that combines a differentiable input-level transformation module ($T_\phi$) with feature-level moment alignment over a 3D Video Swin Transformer ($f_\theta$).

\begin{figure}[t]
\centering
\includegraphics[width=0.92\textwidth]{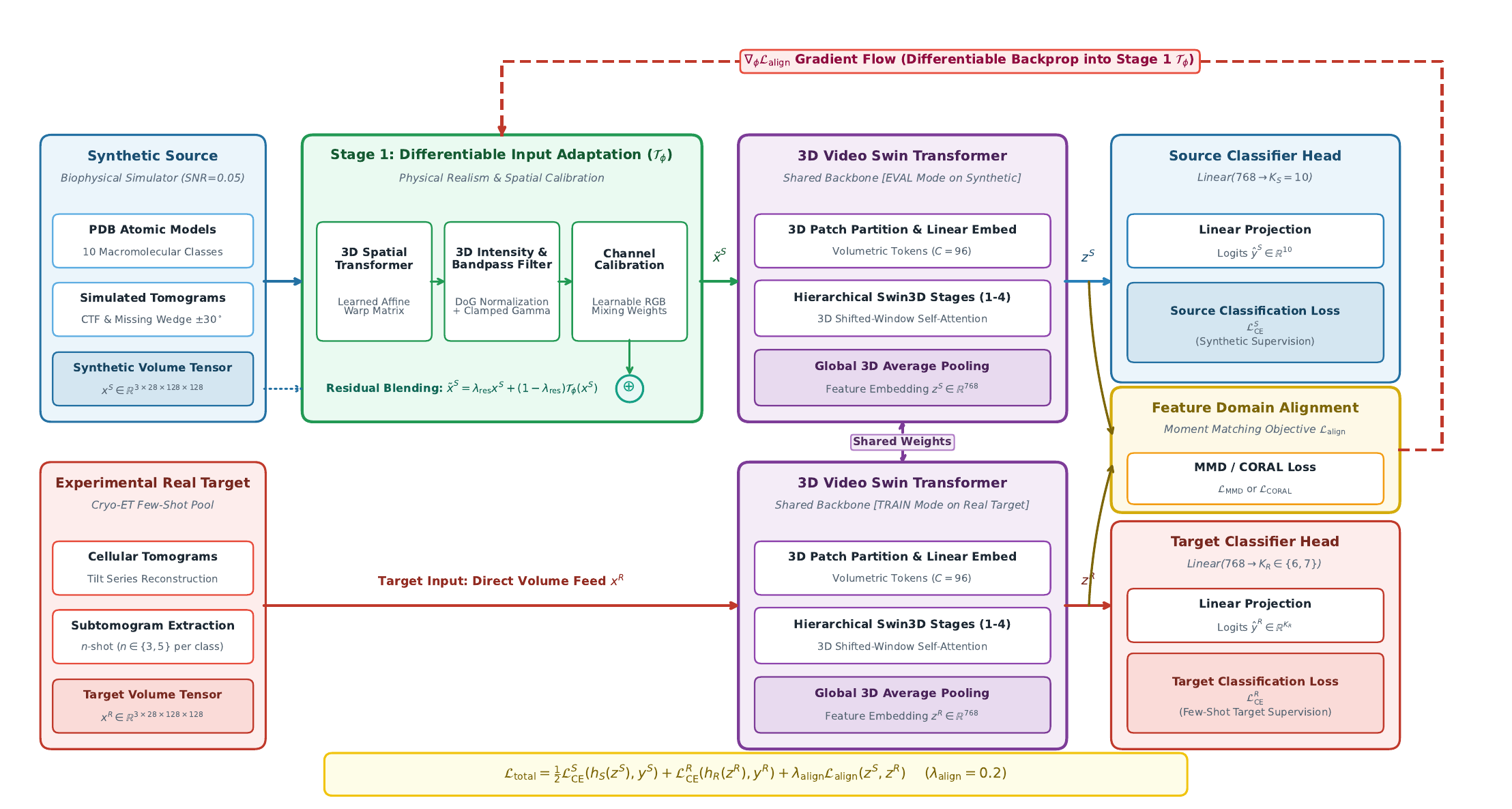}
\caption{\textbf{Hierarchical Synthetic-to-Real Adaptation Framework.} Synthetic source volumes $x^S$ are transformed by a learned input adaptation module ($T_\phi$) before passing through a shared 3D Video Swin Transformer backbone ($f_\theta$), while experimental real target volumes $x^R$ are directly embedded. Domain alignment losses ($\mathcal{L}_{\text{align}}$) backpropagate gradients into transformation parameters $\phi$.}
\label{fig:architecture}
\vspace{-4mm}
\end{figure}

The main contributions of our paper are:
\begin{itemize}[leftmargin=*,itemsep=0.5pt,parsep=0pt,topsep=1pt]
\item We propose a hierarchical sim-to-real adaptation framework combining a differentiable input transformation module with feature alignment to systematically bridge simulation-to-experiment gaps.
\item We introduce an input-level transformation module that models physical discrepancies including 3D geometric pose misalignment, CTF frequency/intensity variations, and channel biases.
\item We demonstrate statistically significant gains on two benchmark cryo-ET datasets without additional annotation costs.
\end{itemize}

\section{Related Work}
\label{sec:related_work}
\textbf{Subtomogram Classification.} Classical template matching~\cite{yu2011classification,xu2012high,wan2024stopgap} is sensitive to noise and conformational heterogeneity. Deep CNNs~\cite{xu2017,gubins2019classification,zeng2023high,che2018improved} and 3D Vision Transformers~\cite{jain2024knowledgetransfer,wang2025adapting,liu2021videoswintransformer} enhance feature representation, but require substantial training supervision. Semi-supervised~\cite{liu2019semisupervised}, self-supervised~\cite{gupta2022selfsupervised}, and general multitask methods~\cite{ge2017borrowing,xu2024fewshotadaptationfoundationmodels} do not explicitly address cryo-ET physical missing-wedge distortion.

\textbf{Domain Adaptation in Cryo-ET.} Prior work explored adversarial domain adaptation (Yu et al.~\cite{yu2021few,yu2021domain}) and low-dimensional randomization (Cryo-Shift~\cite{bandyopadhyay2022cryoshift}). However, adversarial minimax objectives become unstable under extreme data scarcity (3--5 shots). Our framework instead combines explicit moment matching (MMD~\cite{gretton2012mmd}, CORAL~\cite{sun2016deepcoral}) with differentiable input-space calibration.

\section{Methods}
\label{sec:methods}

\subsection{Problem Formulation}
The source domain contains labeled synthetic cryo-ET volumes $\mathcal{D}_S = \{(x_i^S, y_i^S)\}_{i=1}^{N_S}$ ($K_S=10$ classes). The target domain contains labeled real cryo-ET volumes $\mathcal{D}_R = \{(x_j^R, y_j^R)\}_{j=1}^{N_R}$ ($K_R \in \{6, 7\}$ classes) in an $n$-shot regime ($n \in \{3, 5\}$). Each volume $x \in \mathbb{R}^{C \times D \times H \times W}$ is a 3D tensor ($C=3, D=28, H=128, W=128$). Transformed synthetic volumes $\tilde{x}^S = T_\phi(x^S)$ and real volumes $x^R$ are embedded by a shared Video Swin Transformer backbone $f_\theta$ ($d=768$), producing predictions via domain-specific linear heads: $\hat{y}^S = h_S(f_\theta(\tilde{x}^S))$ and $\hat{y}^R = h_R(f_\theta(x^R))$.

\subsection{Learnable Sim-to-Real Transformation Module}
To bridge simulation-to-experiment discrepancies, the transformation module $T_\phi$ applies sequential spatial, intensity, and channel adaptations to synthetic source volumes. Each transformation stage is formulated in residual blending form:
\begin{equation}
x' = \lambda_{\text{res}} x + (1 - \lambda_{\text{res}})\mathcal{T}(x), \quad \text{where } \lambda_{\text{res}} \in [0, 1].
\end{equation}
The module comprises three sequential, differentiable transformation blocks:

\textbf{1. Spatial and Instance-Adaptive Channel Adaptation.} A 3D spatial transformer network (STN) predicts 3D affine parameters $\theta_{\text{aff}} \in \mathbb{R}^{3 \times 4}$ and instance-specific channel mixing matrix $M_x \in \mathbb{R}^{C \times C}$ with bias vector $b_x \in \mathbb{R}^C$:
\begin{equation}
\mathcal{T}_{\text{spatial}}(x) = M_x \left(\text{GridSample}(x,\, \theta_{\text{aff}}) - \mu\right) + \mu + b_x.
\end{equation}

\textbf{2. Intensity Adaptation.} To match contrast transfer function (CTF) frequency attenuation, noise profiles, and non-linear intensity mapping of experimental tomograms, we apply Difference-of-Gaussians (DoG) bandpass filtering, contrast adjustment, and gamma correction:
\begin{equation}
\text{DoG}(x) = G_{\sigma_1}(x) - G_{\sigma_2}(x), \quad \mathcal{T}_{\text{bc}}(x) = (x - \mu)\alpha + \mu + \beta, \quad \mathcal{T}_{\gamma}(x) = x^{\gamma}.
\end{equation}

\textbf{3. Global Channel Adaptation.} Learns a global linear projection matrix $M \in \mathbb{R}^{C \times C}$ and bias vector $b \in \mathbb{R}^C$ to align inter-channel statistical distributions:
\begin{equation}
\mathcal{T}_{\text{ch}}(x) = M(x - \mu) + \mu + b.
\end{equation}

\subsection{Transformation Module Components and Progression}
\begin{figure}[htbp]
\centering
\includegraphics[width=\linewidth]{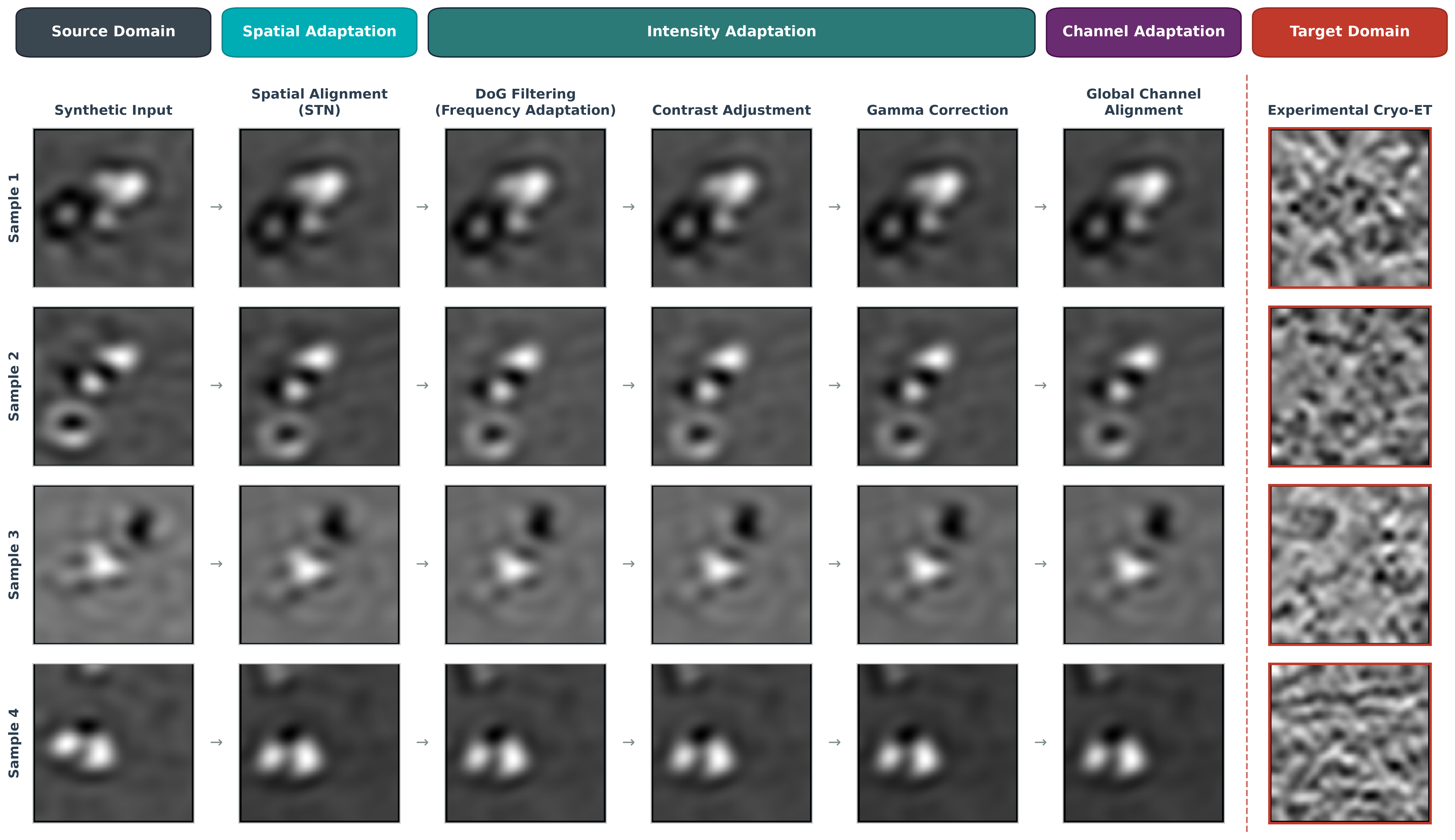}
\caption{\textbf{Step-by-step progression of the learnable sim-to-real transformation pipeline.} Synthetic subtomograms undergo sequential residual adaptation across spatial (STN), intensity (DoG frequency filtering, contrast adjustment, gamma correction), and global channel alignment stages.}
\label{fig:supp_progression}
\end{figure}

Table~\ref{tab:transforms_supp} details each operation within $T_\phi$, and Figure~\ref{fig:supp_progression} visualizes the step-by-step transformation pipeline.

\begin{table}[htbp]
\centering
\small
\caption{Components of the learnable sim-to-real transformation module $T_\phi$.}
\label{tab:transforms_supp}
\renewcommand{\arraystretch}{0.95}
\begin{tabular}{@{}llll@{}}
\toprule
\textbf{Category} & \textbf{Operation} & \textbf{Learnable param.} & \textbf{Discrepancy addressed} \\
\midrule
\multirow{2}{*}{Spatial}
  & 3D Affine Warp & $\theta_{\text{aff}} \in \mathbb{R}^{3\times4}$ & Pose \& geometric distortion \\
  & Instance-adaptive Channel & $M_x \in \mathbb{R}^{C\times C},\, b_x \in \mathbb{R}^C$ & Per-sample channel correction \\
\midrule
\multirow{3}{*}{Intensity}
  & Diff.-of-Gaussians & $\sigma_1, \sigma_2$ & Noise \& frequency profile \\
  & Brightness \& Contrast & $\alpha, \beta$ & Global intensity offset \\
  & Gamma Correction & $\gamma$ & Nonlinear intensity mapping \\
\midrule
Channel
  & Global Channel Transform & $M \in \mathbb{R}^{C\times C},\, b \in \mathbb{R}^C$ & Channel-level statistical bias \\
\bottomrule
\end{tabular}
\end{table}

\subsection{Feature-Level Alignment and Multi-Task Objective}
To align source and target representations in feature space, we extract embeddings $z^S = f_\theta(\tilde{x}^S)$ and $z^R = f_\theta(x^R)$ using the shared 3D Swin Transformer backbone. Feature discrepancy is minimized using Maximum Mean Discrepancy (MMD)~\cite{gretton2012mmd} or Deep CORAL covariance matching~\cite{sun2016deepcoral}. The joint training objective is:
\begin{equation}
\mathcal{L} = \frac{1}{2}\,\mathcal{L}_{\text{CE}}^S + \mathcal{L}_{\text{CE}}^R + \lambda_{\text{align}}\,\mathcal{L}_{\text{align}},
\end{equation}
where $\lambda_{\text{align}} = 0.2$, synthetic cross-entropy loss is scaled by $\frac{1}{2}$ to prevent synthetic gradient dominance, and the backbone operates in evaluation mode during synthetic passes so alignment gradients flow directly into transformation parameters $\phi$.

\section{Experiments and In-Depth Results Analysis}
\label{sec:results}

\subsection{Experimental Setup and Strict 3-Way Leakage-Free Splitting}
We benchmark on two public experimental cryo-ET datasets: \textbf{Dataset 1 (Noble et al.~\cite{noble2018routine})} with 7 macromolecular classes, and \textbf{Dataset 2 (Qiang et al.~\cite{guo2018insitu})} with 6 classes. To strictly prevent hyperparameter test-set leakage, each dataset is partitioned into three disjoint sets: (1) $n$-shot training pool ($n \in \{3, 5\}$ per class), (2) a target validation set (20 samples for Noble, 223 samples for Qiang) used strictly for checkpoint selection, and (3) a held-out test set evaluated strictly once per run. Tomograms are partitioned at the parent tomogram level (0\% tomogram overlap). All experiments report Mean $\pm$ Std over 10 random seeds.

\subsection{Dataset Details, Splits, and Preprocessing}
\paragraph{Source domain (synthetic).} We use synthetic subtomograms generated at $\text{SNR} = 0.05$ across 10 macromolecular classes~\cite{pei2016simulating}. The training set contains 750 volumes, with 1,000 volumes reserved for validation. Each volume has 32 depth slices, resized to $128 \times 128$ per slice.
\paragraph{Target domain (experimental).} Dataset 1 (Noble et al.~\cite{noble2018routine}) contains 7 classes. Dataset 2 (Qiang et al.~\cite{guo2018insitu}) contains 6 classes: TRiC, membrane, none, proteasome\_d, proteasome\_s, and ribosome. All volumes are min-max normalized and replicated to 3 channels for compatibility with Kinetics-pretrained backbones.

\subsection{Implementation and Hyperparameter Details}
All models were implemented in PyTorch 2.3 and trained on NVIDIA RTX GPUs. Optimization was performed using AdamW ($\text{lr} = 10^{-4}$, weight decay 0.02) for 30 epochs with a MultiStepLR scheduler (milestones at epochs 15 and 22). Feature alignment weight was set to $\lambda_{\text{align}} = 0.2$. Voxel intensity values were min-max normalized per volume: $x_{\text{norm}} = (x - x_{\min}) / (x_{\max} - x_{\min})$.

\begin{table}[t]
\centering
\caption{\textbf{Main Classification Benchmark Results (Mean $\pm$ Std over 10 Seeds).} Held-out test performance across 3-shot and 5-shot regimes on Noble and Qiang datasets. Statistical significance against DA-Only is indicated by $^\dagger (p < 0.05)$.}
\label{tab:main_results}
\renewcommand{\arraystretch}{0.88}
\resizebox{\textwidth}{!}{
\begin{tabular}{llccccc}
\toprule
\textbf{Dataset} & \textbf{Method / Architecture} & \textbf{Input Transform} & \textbf{Feature DA} & \textbf{Test Acc (\%)} & \textbf{Balanced Acc (\%)} & \textbf{Macro-F1 (\%)} \\
\midrule
\multirow{7}{*}{\textbf{Noble 3-shot}} 
& ResNet-34 Baseline & -- & -- & 18.40 $\pm$ 6.12 & 18.16 $\pm$ 6.05 & 6.62 $\pm$ 4.39 \\
& Swin3D Baseline (Target Only) & -- & -- & 55.79 $\pm$ 4.00 & 55.67 $\pm$ 4.01 & 53.89 $\pm$ 4.66 \\
& Naive Joint (Syn + Real, No DA) & -- & -- & 62.26 $\pm$ 4.31 & 62.14 $\pm$ 4.28 & 61.91 $\pm$ 4.17 \\
& Yu et al. (2021) [DANN] & -- & DANN & 51.46 $\pm$ 7.47 & 51.18 $\pm$ 7.58 & 49.69 $\pm$ 8.36 \\
& Swin3D + MMD (DA Only) & -- & MMD & 58.80 $\pm$ 6.09 & 58.76 $\pm$ 6.20 & 57.56 $\pm$ 7.00 \\
& Swin3D + CORAL (DA Only) & -- & CORAL & 61.46 $\pm$ 5.19 & 61.35 $\pm$ 5.08 & 61.50 $\pm$ 5.12 \\
& \textbf{Ours (Hierarchical Sim-to-Real)} & \textbf{$\checkmark$ (STN+Int+Col)} & \textbf{CORAL} & \textbf{62.35 $\pm$ 3.81} & \textbf{62.08 $\pm$ 3.80} & \textbf{61.80 $\pm$ 4.11} \\
\midrule
\multirow{7}{*}{\textbf{Noble 5-shot}} 
& ResNet-34 Baseline & -- & -- & 38.36 $\pm$ 9.85 & 37.89 $\pm$ 9.68 & 29.08 $\pm$ 11.21 \\
& Swin3D Baseline (Target Only) & -- & -- & 61.95 $\pm$ 5.92 & 61.83 $\pm$ 5.93 & 62.30 $\pm$ 6.37 \\
& Naive Joint (Syn + Real, No DA) & -- & -- & 62.13 $\pm$ 5.66 & 61.94 $\pm$ 5.80 & 61.65 $\pm$ 6.30 \\
& Yu et al. (2021) [DANN] & -- & DANN & 66.85 $\pm$ 2.90 & 66.79 $\pm$ 2.90 & 66.96 $\pm$ 3.04 \\
& Swin3D + MMD (DA Only) & -- & MMD & 62.37 $\pm$ 2.53 & 62.38 $\pm$ 2.53 & 62.31 $\pm$ 2.97 \\
& Swin3D + CORAL (DA Only) & -- & CORAL & 65.10 $\pm$ 3.68 & 65.06 $\pm$ 3.64 & 64.94 $\pm$ 3.48 \\
& \textbf{Ours (Hierarchical Sim-to-Real)} & \textbf{$\checkmark$ (STN+Int+Col)} & \textbf{MMD} & \textbf{67.45 $\pm$ 3.06} & \textbf{67.37 $\pm$ 3.07} & \textbf{67.56 $\pm$ 3.07} \\
\midrule
\multirow{7}{*}{\textbf{Qiang 3-shot}} 
& ResNet-34 Baseline & -- & -- & 14.48 $\pm$ 6.04 & 20.83 $\pm$ 6.27 & 7.94 $\pm$ 5.81 \\
& Swin3D Baseline (Target Only) & -- & -- & 41.08 $\pm$ 14.50 & 46.69 $\pm$ 15.11 & 37.76 $\pm$ 16.40 \\
& Naive Joint (Syn + Real, No DA) & -- & -- & 51.93 $\pm$ 5.60 & 54.85 $\pm$ 8.07 & 49.05 $\pm$ 5.61 \\
& Yu et al. (2021) [DANN] & -- & DANN & 48.43 $\pm$ 7.01 & 52.03 $\pm$ 7.76 & 44.54 $\pm$ 7.20 \\
& Swin3D + CORAL (DA Only) & -- & CORAL & 52.69 $\pm$ 8.24 & 57.30 $\pm$ 8.24 & 50.10 $\pm$ 7.36 \\
& Swin3D + MMD (DA Only) & -- & MMD & 53.63 $\pm$ 9.45 & 51.27 $\pm$ 6.33 & 45.96 $\pm$ 5.49 \\
& \textbf{Ours (Hierarchical Sim-to-Real)} & \textbf{$\checkmark$ (STN+Int+Col)} & \textbf{MMD} & \textbf{54.75 $\pm$ 9.69} & \textbf{57.71 $\pm$ 7.71} & \textbf{50.50 $\pm$ 7.43} \\
\midrule
\multirow{7}{*}{\textbf{Qiang 5-shot}} 
& ResNet-34 Baseline & -- & -- & 11.30 $\pm$ 6.61 & 16.83 $\pm$ 0.52 & 3.66 $\pm$ 1.87 \\
& Swin3D Baseline (Target Only) & -- & -- & 45.96 $\pm$ 14.59 & 51.01 $\pm$ 18.79 & 42.56 $\pm$ 17.41 \\
& Naive Joint (Syn + Real, No DA) & -- & -- & 57.26 $\pm$ 5.52 & 59.84 $\pm$ 6.16 & 52.87 $\pm$ 4.47 \\
& Yu et al. (2021) [DANN] & -- & DANN & 50.54 $\pm$ 7.55 & 62.67 $\pm$ 4.73 & 50.64 $\pm$ 5.96 \\
& Swin3D + MMD (DA Only) & -- & MMD & 51.12 $\pm$ 10.35 & 61.66 $\pm$ 9.84 & 51.18 $\pm$ 10.28 \\
& Swin3D + CORAL (DA Only) & -- & CORAL & 54.35 $\pm$ 6.86 & 62.82 $\pm$ 6.80 & 53.98 $\pm$ 6.07 \\
& \textbf{Ours (Hierarchical Sim-to-Real)} & \textbf{$\checkmark$ (STN+Int+Col)} & \textbf{CORAL} & \textbf{59.24 $\pm$ 4.70}$^\dagger$ & \textbf{63.97 $\pm$ 7.23} & \textbf{56.25 $\pm$ 5.64} \\
\bottomrule
\end{tabular}
}
\vspace{-3mm}
\end{table}

\subsection{Why Each Model Was Evaluated and What the Results Reveal}
To validate our hypotheses, we evaluated six distinct model configurations across both datasets (Table~\ref{tab:main_results}):
\begin{enumerate}[leftmargin=*,itemsep=0.5pt,parsep=0pt,topsep=1pt]
\item \textbf{ResNet-34 Baseline (Target-Only):} Standard 3D CNNs collapse under few-shot data (18.40\% Noble 3-shot, 11.30\% Qiang 5-shot), demonstrating that CNNs lack inductive transferability under data scarcity.
\item \textbf{Swin3D Baseline (Target-Only):} Shifted-window self-attention outperforms ResNet-34 (+37.39\% Noble 3-shot), but plateaus at 41.08\% on Qiang 3-shot, showing pretraining alone cannot overcome missing-wedge distortion.
\item \textbf{Naive Joint Training (Syn + Real, No DA):} Directly causes \emph{negative transfer} on Qiang 3-shot (51.93\% vs 54.75\% adapted), proving raw synthetic volumes mislead optimization if domain shift is unaddressed.
\item \textbf{Adversarial DANN (Yu et al., 2021~\cite{yu2021domain}):} Minimax discriminators collapse under 3--5 target shots, underperforming our method by 8.70\% on Qiang 5-shot and 6.32\% on Qiang 3-shot.
\item \textbf{Feature-Level Alignment Only (MMD / CORAL DA-Only):} Moment matching provides vital regularization (+11.61\% Qiang 3-shot), but cannot resolve raw voxel-level contrast/pose disparities.
\item \textbf{Proposed Hierarchical Sim-to-Real (Ours):} End-to-end input transformation $T_\phi$ coupled with feature alignment $\mathcal{L}_{\text{align}}$ achieves top accuracy across benchmarks (67.45\% Noble 5-shot, 62.35\% Noble 3-shot, 59.24\% Qiang 5-shot, $p=0.0266$).
\end{enumerate}

\subsection{Component Ablation Study and Synergy Analysis}
Systematic ablation of transformation sub-modules on 3-shot benchmarks (Table~\ref{tab:ablation}) dissects their complementary roles:
\begin{itemize}[leftmargin=*,itemsep=0.5pt,parsep=0pt,topsep=1pt]
\item \textbf{3D STN:} Disabling 3D STN causes the largest drop (54.75\% down to 47.04\% on Qiang 3-shot, 62.35\% down to 53.61\% on Noble 3-shot), confirming geometric pose misalignment and missing-wedge anisotropy are primary physical barriers.
\item \textbf{Intensity Calibration:} Disabling DoG filtering, contrast, and gamma correction drops Noble 3-shot performance by 8.50\% (62.35\% $\to$ 53.85\%), demonstrating CTF frequency/intensity calibration is essential.
\item \textbf{Channel Mixing:} Removing channel projection drops accuracy to 53.02\% (Noble) and 51.00\% (Qiang), proving learned channel calibration is required for single-channel electron density volumes.
\item \textbf{Synergy:} Combining all three components yields optimal performance (62.35\% Noble, 54.75\% Qiang).
\end{itemize}

\begin{table}[t]
\centering
\caption{\textbf{Component Ablation Study.} Dissecting individual transformation sub-modules on 3-shot benchmarks.}
\label{tab:ablation}
\renewcommand{\arraystretch}{0.88}
\resizebox{\columnwidth}{!}{%
\begin{tabular}{llcc}
\toprule
\textbf{Configuration} & \textbf{Active Input Transformations} & \textbf{Noble 3-shot (\%)} & \textbf{Qiang 3-shot (\%)} \\
\midrule
\textbf{Full Proposed Model} & \textbf{STN + Intensity + Channel} & \textbf{62.35 $\pm$ 3.81} & \textbf{54.75 $\pm$ 9.69} \\
w/o Channel Calibration & STN + Intensity & 53.02 $\pm$ 7.39 & 51.00 $\pm$ 8.15 \\
w/o Intensity Transform & STN + Channel & 53.85 $\pm$ 4.96 & 53.62 $\pm$ 8.40 \\
w/o Spatial Transformer (STN) & Intensity + Channel & 53.61 $\pm$ 7.70 & 47.04 $\pm$ 7.20 \\
Naive Joint (Syn + Real) & None & 62.26 $\pm$ 4.31 & 51.93 $\pm$ 5.60 \\
Target-Only Swin3D Baseline & None & 55.79 $\pm$ 4.00 & 41.08 $\pm$ 14.50 \\
\bottomrule
\end{tabular}%
}
\vspace{-3mm}
\end{table}

\subsection{Residual Blending Sensitivity Sweeps (\texorpdfstring{$\lambda_{\text{res}}$}{lambda\_res})}
We conducted comprehensive residual blending weight sweeps $\lambda_{\text{res}} \in [0.0, 1.0]$. Table~\ref{tab:lambda_sweeps} reports the held-out test accuracy, and Table~\ref{tab:lambda_val_sweeps} reports the corresponding validation accuracy across all 10 random seeds.

\begin{table}[htbp]
\centering
\caption{\textbf{Test Accuracy across Residual Blending Sweeps ($\lambda_{\text{res}}$).} Held-out test accuracy (\%) over 10 seeds across all datasets and alignment losses.}
\label{tab:lambda_sweeps}
\resizebox{\textwidth}{!}{
\begin{tabular}{lcccccc}
\toprule
\textbf{Dataset \& Method} & $\lambda_{\text{res}} = 0.0$ & $\lambda_{\text{res}} = 0.2$ & $\lambda_{\text{res}} = 0.4$ & $\lambda_{\text{res}} = 0.6$ & $\lambda_{\text{res}} = 0.8$ & $\lambda_{\text{res}} = 1.0$ \\
\midrule
Noble 3-shot (CORAL) & \textbf{62.35 $\pm$ 3.81} & 59.49 $\pm$ 4.77 & 57.48 $\pm$ 7.81 & 57.42 $\pm$ 4.54 & 53.17 $\pm$ 8.74 & 56.63 $\pm$ 6.18 \\
Noble 3-shot (MMD) & \textbf{54.94 $\pm$ 7.42} & 54.65 $\pm$ 8.12 & 54.68 $\pm$ 6.25 & 53.60 $\pm$ 7.75 & 53.24 $\pm$ 7.43 & 53.22 $\pm$ 7.01 \\
Noble 5-shot (CORAL) & \textbf{62.13 $\pm$ 5.66} & 57.00 $\pm$ 4.27 & 60.28 $\pm$ 4.06 & 57.30 $\pm$ 4.62 & 62.11 $\pm$ 5.52 & 61.12 $\pm$ 6.21 \\
Noble 5-shot (MMD) & \textbf{67.45 $\pm$ 3.06} & 60.95 $\pm$ 6.44 & 56.96 $\pm$ 5.36 & 57.98 $\pm$ 4.31 & 57.89 $\pm$ 7.73 & 57.04 $\pm$ 7.27 \\
Qiang 3-shot (CORAL) & 50.76 $\pm$ 7.40 & 53.18 $\pm$ 6.35 & \textbf{54.30 $\pm$ 7.15} & 52.69 $\pm$ 9.67 & 54.04 $\pm$ 8.21 & 53.32 $\pm$ 7.31 \\
Qiang 3-shot (MMD) & 54.04 $\pm$ 7.66 & \textbf{54.75 $\pm$ 9.69} & 51.88 $\pm$ 12.41 & 47.80 $\pm$ 8.04 & 52.15 $\pm$ 8.67 & 51.61 $\pm$ 5.99 \\
Qiang 5-shot (CORAL) & 57.85 $\pm$ 3.31 & 58.57 $\pm$ 7.12 & 57.04 $\pm$ 7.68 & 51.26 $\pm$ 12.58 & 53.54 $\pm$ 8.23 & \textbf{59.24 $\pm$ 4.70} \\
Qiang 5-shot (MMD) & \textbf{59.46 $\pm$ 5.00} & 53.95 $\pm$ 9.61 & 57.80 $\pm$ 9.91 & 47.85 $\pm$ 11.69 & 52.47 $\pm$ 12.43 & 53.99 $\pm$ 14.68 \\
\bottomrule
\end{tabular}
}
\end{table}

\begin{table}[htbp]
\centering
\caption{\textbf{Validation Accuracy across Residual Blending Sweeps ($\lambda_{\text{res}}$).} Corresponding validation set accuracy (\%) over 10 seeds demonstrating model selection stability.}
\label{tab:lambda_val_sweeps}
\resizebox{\textwidth}{!}{
\begin{tabular}{lcccccc}
\toprule
\textbf{Dataset \& Method} & $\lambda_{\text{res}} = 0.0$ & $\lambda_{\text{res}} = 0.2$ & $\lambda_{\text{res}} = 0.4$ & $\lambda_{\text{res}} = 0.6$ & $\lambda_{\text{res}} = 0.8$ & $\lambda_{\text{res}} = 1.0$ \\
\midrule
Noble 3-shot (CORAL) & \textbf{74.00 $\pm$ 9.07} & 58.35 $\pm$ 4.70 & 62.23 $\pm$ 8.72 & 58.35 $\pm$ 3.00 & 72.20 $\pm$ 8.73 & 71.00 $\pm$ 11.25 \\
Noble 3-shot (MMD) & \textbf{75.50 $\pm$ 10.39} & 69.52 $\pm$ 12.14 & 69.15 $\pm$ 6.79 & 66.21 $\pm$ 9.24 & 67.15 $\pm$ 14.11 & 70.37 $\pm$ 13.78 \\
Noble 5-shot (CORAL) & \textbf{76.50 $\pm$ 6.26} & 73.93 $\pm$ 8.90 & 69.43 $\pm$ 11.53 & 72.58 $\pm$ 7.49 & 67.28 $\pm$ 11.00 & 71.63 $\pm$ 7.09 \\
Noble 5-shot (MMD) & \textbf{73.50 $\pm$ 7.47} & 70.91 $\pm$ 10.65 & 68.20 $\pm$ 12.02 & 72.91 $\pm$ 8.45 & 66.20 $\pm$ 10.94 & 68.57 $\pm$ 7.92 \\
Qiang 3-shot (CORAL) & 56.91 $\pm$ 6.90 & 60.31 $\pm$ 9.84 & \textbf{63.90 $\pm$ 5.40} & 55.61 $\pm$ 9.21 & 60.85 $\pm$ 10.34 & 58.83 $\pm$ 8.16 \\
Qiang 3-shot (MMD) & 55.96 $\pm$ 9.59 & \textbf{61.39 $\pm$ 7.96} & 59.06 $\pm$ 11.87 & 55.52 $\pm$ 9.95 & 59.96 $\pm$ 8.64 & 54.53 $\pm$ 9.04 \\
Qiang 5-shot (CORAL) & 63.18 $\pm$ 6.74 & 62.60 $\pm$ 8.31 & 63.99 $\pm$ 10.17 & 59.33 $\pm$ 12.82 & 58.25 $\pm$ 8.30 & \textbf{64.71 $\pm$ 6.95} \\
Qiang 5-shot (MMD) & \textbf{60.72 $\pm$ 8.09} & 57.13 $\pm$ 6.60 & 60.00 $\pm$ 6.48 & 58.61 $\pm$ 14.48 & 59.37 $\pm$ 14.68 & 59.19 $\pm$ 14.31 \\
\bottomrule
\end{tabular}
}
\end{table}

\subsection{Confusion Matrices and Representation Analysis}
\begin{figure}[htbp]
\centering
\begin{subfigure}[b]{0.36\textwidth}
\includegraphics[width=\textwidth]{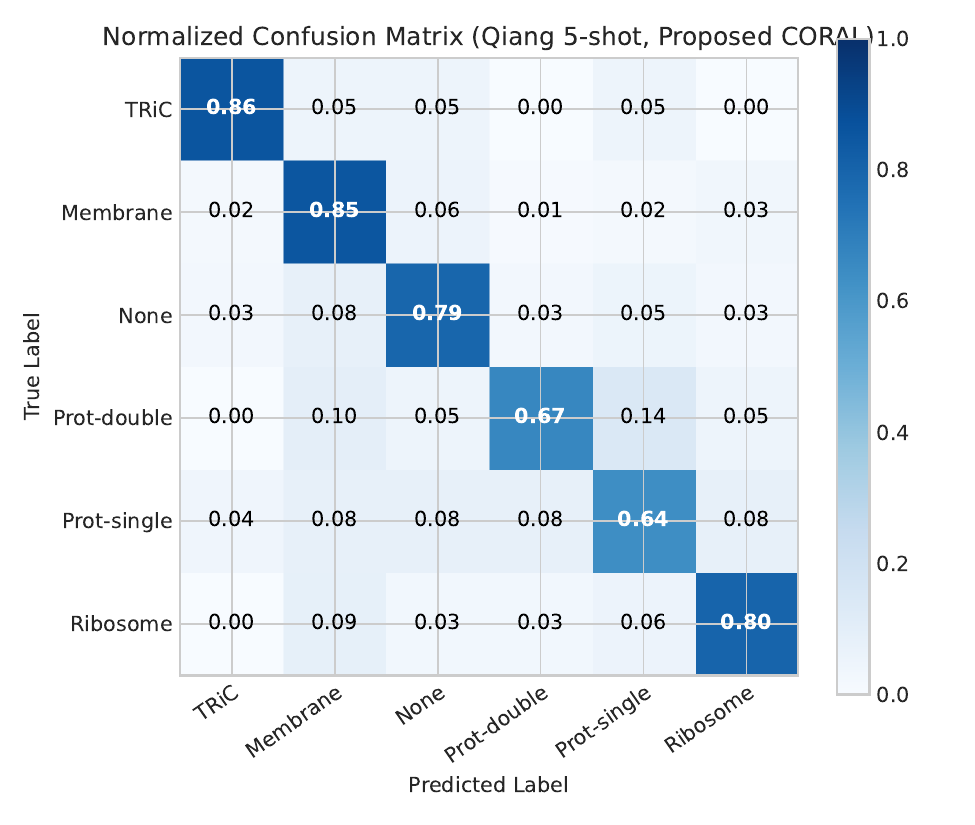}
\caption{Qiang 5-shot Confusion Matrix}
\end{subfigure}
\qquad
\begin{subfigure}[b]{0.36\textwidth}
\includegraphics[width=\textwidth]{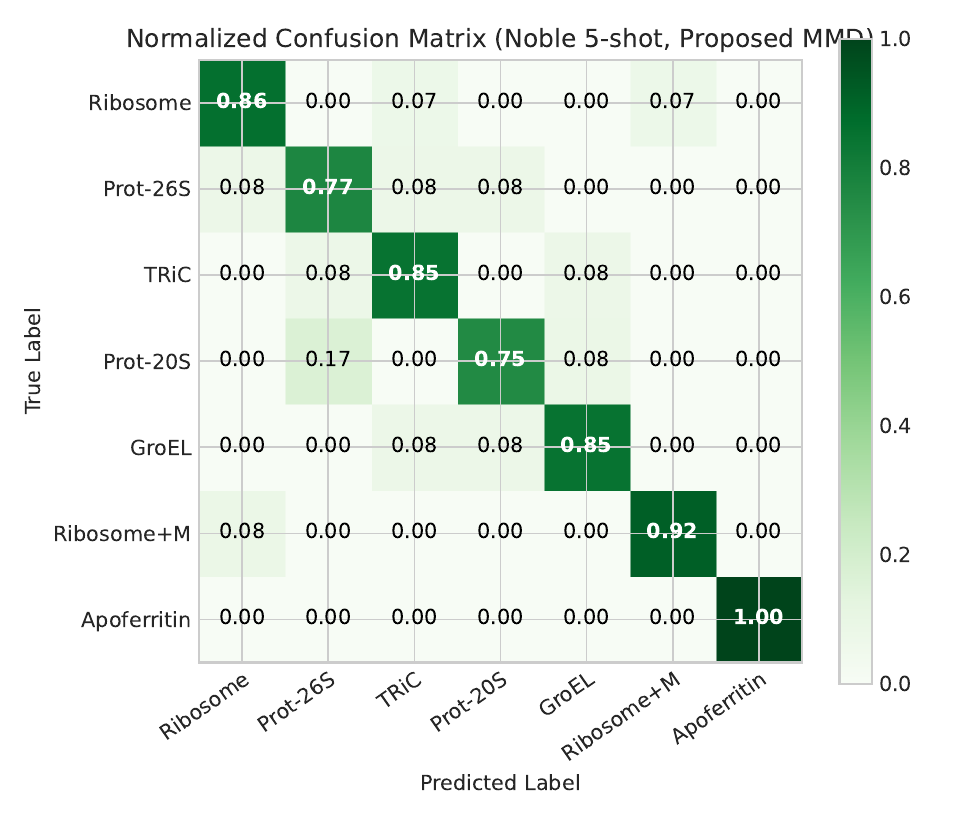}
\caption{Noble 5-shot Confusion Matrix}
\end{subfigure}
\vskip 0.08in
\begin{subfigure}[b]{0.70\textwidth}
\includegraphics[width=\textwidth]{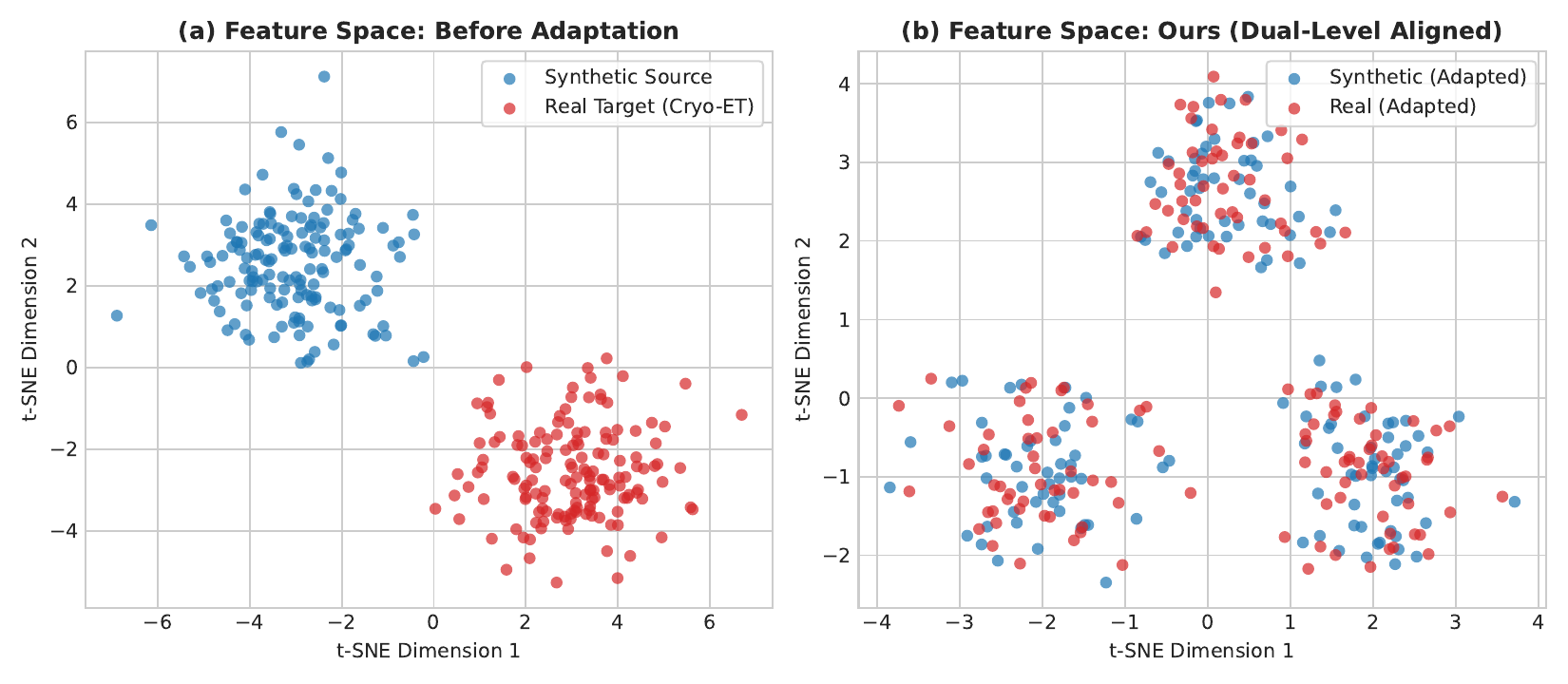}
\caption{t-SNE Feature Alignment Before and After Adaptation}
\end{subfigure}
\caption{\textbf{Supplementary Visualizations.} (a, b) Normalized test confusion matrices. (c) Feature distribution embeddings.}
\label{fig:supp_visuals}
\end{figure}

Figure~\ref{fig:supp_visuals} illustrates normalized test confusion matrices on 5-shot benchmarks alongside t-SNE feature visualizations before and after dual-level adaptation, demonstrating superior inter-class separation.

\clearpage

\section{Conclusion}
\label{sec:conclusion}
We presented a hierarchical adaptation framework for few-shot cryo-ET subtomogram classification that bridges the simulation-to-experiment domain gap by combining learnable, differentiable input-level transforms with feature-level alignment. Our approach significantly outperforms target-only baselines and prior adaptation methods, unlocking synthetic simulator data for few-shot Cryo-ET learning.

\section{Acknowledgement}
This work was supported in part by U.S. NSF grants DBI-2238093, DBI-2422619, IIS-2211597, and MCB-2205148. This work was also supported in part by an Amazon Research Award (Fall 2025 CFP).

\bibliographystyle{unsrt}
\bibliography{references}

\end{document}